%% file: main.tex
\documentclass{article}

\usepackage{authblk}
\usepackage[backend=biber]{biblatex}
\usepackage{arxiv}
\usepackage{hyperref}

\hypersetup{
  pdftitle={A Survey of Active Learning for Text Classification using Deep Neural Networks},
  pdfsubject={cs.CL},
  pdfauthor={Christopher Schr\"oder, Andreas Niekler},
  pdfkeywords={active learning, neural networks, text classification},
}

\usepackage{times}
\usepackage{url}
\usepackage{latexsym}
 
\usepackage[utf8]{inputenc} 
\usepackage[T1]{fontenc}    
\usepackage{url}            
\usepackage{booktabs}       
\usepackage{amsfonts}       
\usepackage{nicefrac}       
\usepackage{microtype}      
\usepackage[x11names]{xcolor}
\usepackage{graphicx}

\usepackage{array}
\newcolumntype{P}[1]{>{\raggedright\arraybackslash}p{#1}}

\usepackage{framed}
\usepackage{forest} 
 
\usepackage{longtable}
\usepackage[tableposition=below, aboveskip=10pt]{caption}
\usetikzlibrary{arrows.meta,shapes,positioning,shadows,trees,decorations.pathreplacing}
\ProvidesForestLibrary{edges}

\tikzset{
    basic/.style  = {draw, text width=2cm, drop shadow, font=\sffamily, rectangle},
    root/.style   = {basic, thin,
                     fill=white, text width=2.6cm},
    onode/.style = {basic, thin, fill=white,text width=2.6cm,},
    onode_dashed/.style = {onode, dashed},
    tnode/.style = {basic, thin, align=left, fill=white, text width=10.5em},
    tnode_nlp/.style = {tnode, double},
    edge from parent/.style={draw=black, edge from parent fork right},
    zlevel/.style={%
    execute at begin scope={\pgfonlayer{#1}},
    execute at end scope={\endpgfonlayer}
}
} 

\tikzset{zlevel/.style={%
    execute at begin scope={\pgfonlayer{#1}},
    execute at end scope={\endpgfonlayer}
}}

\forestset{ 
    declare dimen={fork sep}{0.5em},
    forked edge/.style={
      edge={rotate/.pgfmath=grow()},
      edge path'={(!u.parent anchor) -- ++(\forestoption{fork sep},0) |- (.child anchor)},
    },
    forked edges/.style={
      for tree={parent anchor=children},
      for descendants={child anchor=parent,forked edge}
    }
} 
 
\newcommand{\pindent}{\hspace*{1em}}

\newcommand{\phrasing}[1]{{#1}}

\newcommand{\revised}[1]{{#1}}

\newcommand{\hiddennotes}[1]{}

\title{A Survey of Active Learning for Text Classification using Deep Neural Networks}

\author{\textbf{Christopher Schröder}}
\author{\textbf{Andreas Niekler}}

\affil{Natural Language Processing Group, University of Leipzig\vspace*{-0.8em}}
\affil{\texttt{cschroeder@uni-leipzig.de}\vspace*{-0.8em}}
\affil{\texttt{andreas.niekler@uni-leipzig.de}}
\date{}

\begin{document}
\maketitle

\begin{abstract}
Natural language processing (NLP) and neural networks (NNs) have both undergone significant changes in recent years.
For active learning (AL) purposes, NNs are, however, less commonly used -- despite their current popularity.
By using the superior text classification performance of NNs for AL, we can either increase a model's performance using the same amount of data or reduce the data and therefore the required annotation efforts while keeping the same performance.
%
%
We review AL for text classification using deep neural networks (DNNs) and elaborate on two main causes which used to hinder the adoption: (a) the inability of NNs to provide reliable uncertainty estimates, on which the most commonly used query strategies rely, and (b) the challenge of training DNNs on small data.
To investigate the former, we construct a taxonomy of query strategies, which distinguishes between data-based, model-based, and prediction-based instance selection, and investigate the prevalence of these classes in recent research.
%
Moreover, we review recent NN-based advances in NLP like word embeddings or language models in the context of (D)NNs, survey the current state-of-the-art at the intersection of AL, text classification, and DNNs and relate recent advances in NLP to AL.
Finally, we analyze recent work in AL for text classification, connect the respective query strategies to the taxonomy, and outline commonalities and shortcomings.
As a result, we highlight gaps in current research and present open research questions.
\end{abstract}

\input{tex/introduction.tex}
\input{tex/related_work.tex}

\input{tex/active_learning.tex}
\input{tex/text_classification.tex}
\input{tex/open_research_questions.tex}

\input{tex/conclusions.tex}

\input{acknowledgements.tex}

\printbibliography

\input{tex/appendix.tex}

\end{document}

%% file: tex/introduction.tex
\section{Introduction}
\label{sec:introduction}


%
%

Data is the fuel of machine learning applications and therefore has been steadily increasing in value.
In many settings an abundant amount of unlabeled data is produced,
but in order to use such data in supervised machine learning, 
one has no choice but to provide labels.
This usually entails a manual labeling process, which is often non-trivial and can even require a domain expert, 
e.g., in patent classification \cite{larkey1999patent,fall2003automated}, or clinical text classification \cite{pestian2007shared,figueroa2012active,garla2013semi-supervised}.
Moreover, this is time-consuming and rapidly increases monetary costs, thereby quickly rendering this approach infeasible.
Even if an expert is available, it is often impossible to label each datum due to the vast size of modern datasets.
This especially impedes the field of Natural Language Processing (NLP), 
in which both the dataset and the amount of text within each document can be huge,
resulting in unbearable amounts of annotation efforts for human experts.
 
%
%
Active Learning (AL) aims to reduce the amount of data annotated by the human expert.
\revised{It is an iterative cyclic process between an {\em oracle} (usually the human annotator)
and an {\em active learner}.}
In contrast to passive learning, in which the data is simply fed to the algorithm, 
the active learner chooses which samples are to be labeled next.  
The labeling itself, however, is done by a human expert, the so-called human in the loop.
Having received new labels, the active learner trains a new model and the process starts from the beginning.
\revised{Using the term active learner, we refer to the composition of a {\em model}, a {\em query strategy}, and a {\em stopping criterion}.}
In this work the model is w.l.o.g. a text classification model, 
the query strategy decides which instances should be labeled next, and the stopping criterion defines when to stop the AL loop.
According to \textcite{settles2010active} there are three main scenarios for AL:
(1) Pool-based, in which the learner has access to the closed set of unlabeled instances, called the pool;
(2) stream-based, where the learner receives one instance at a time and has the options to keep it, or to discard;
(3) membership query synthesis, \revised{in which the learner creates new artificial instances to be labeled}.
If the pool-based scenario operates not on a single instance, but on a batch of instances, this is called {\em batch-mode} AL \cite{settles2010active}.
Throughout this work we assume a pool-based batch-mode scenario because in a text classification setting the dataset is usually a closed set, and the batch-wise operation reduces the number of retraining operations, which cause waiting periods for the user.\\
\pindent \revised{The underlying idea of AL is that few representative instances can be used as surrogate for the full dataset.}
Not only does a smaller subset of the data reduce the computational costs, 
but also it has been shown that AL can even increase the quality of the resulting model compared to learning on the full dataset \cite{schohn2000less,figueroa2012active}.
As a consequence, AL has been used in many NLP tasks, e.g. text classification \cite{tong2001support,hoi2006large-scale}, named entity recognition \cite{shen2004multi-criteria-based,tomanek2009reducing,shen2018deep}, 
or machine translation \cite{haffari2009active} and is still an active area of research.
%


%
%

%
%

%
%

In recent years, deep learning (DL) approaches have dominated most NLP tasks' state-of-the-art results.
This can be attributed to advances in neural networks (NNs), above all Convolutional Neural Networks (CNN; \cite{kim2014convolutional}) and (Bidirectional-)Long Short-Term Memory (LSTM; \cite{hochreiter1997long,graves2005framewise}),
which were eventually adopted into the NLP domain,
and to the advances of using word embeddings \cite{mikolov2013efficient,mikolov2014distributed,pennington2014glove} and contextualized word embeddings \cite{peters2018deep,devlin2019bert}.
Both NN architectures and text representations have raised the state-of-the-art results in the field of text classification considerably (e.g., \cite{zhang2016character-level,howard2018universal,yang2019xlnet}).
If these improvements were transferrable to AL, this would result in a huge increase in efficiency.
For the AL practitioner, this either means achieving the same performance using fewer samples, or having an increase in performance using the same amount of data.
Another favorable development is that transfer learning, especially the paradigm of fine-tuning pre-trained language models (LMs), has become popular in NLP.
In the context of AL this helps especially in the small data scenario, in which a pre-trained model can be leveraged to train a model by fine-tuning using only little data, which would otherwise be infeasible.
Finally, by operating on sub-word units LMs also handle out-of-vocabulary tokens, which is an advantage over many traditional methods.

%
%


%
%

Resulting from these advances, existing AL surveys have become both incomplete in some parts and outdated in others:
They lack comparison against the current state of the art models, do not provide results for more recent large-scale datasets, 
and most importantly, they are lacking the aforementioned advances in NNs and text representations.
Surprisingly, despite the current popularity of NNs, there is only little research about NN-based active learning in the context of NLP, and even less thereof in the context of text classification (see Section \ref{subsect:nn_al} and Section \ref{subsect:active_tc} for a detailed summary).
We suspect this is due to the following reasons: 
(1) Many DL models are known to require large amounts of data \cite{zhang2016character-level},
which is in strong contrast to AL aiming at requiring as little data as possible
(2) there is a whole AL scenario based on artificial data generation, which unfortunately is a lot more challenging for text in contrast to for example images, for which data augmentation is commonly used in classification tasks \cite{wang2017effectiveness};
(3) NNs are lacking uncertainty information regarding their predictions \revised{(as explained in Section \ref{subsect:nn_al})}, which complicates the use of a whole prominent class of query strategies.

\hiddennotes{
  \begin{itemize}
    \item No clear evaluation protocol / huge differences in the evaluation approaches
  \end{itemize}
}


%
%

This survey aims at summarizing the existing approaches of (D)NN-based AL for text classification.
Our main contributions are as follows:

\begin{enumerate}
  \item We provide a taxonomy of query strategies and classify strategies relevant for AL for text classification.
  \item We survey existing work at the intersection of AL, text classification, and (D)NNs.
  \item Recent advances in text classification are summarized and related to the AL process. It is then investigated, if and to what degree they have been adopted for AL.
  \item The experimental setup of previous research is collectively analyzed regarding datasets, models, and query strategies in order to identify recent trends, commonalities, and shortcomings in the experiments.
  \item \revised{We identify research gaps and outline future research directions.}
\end{enumerate}

\noindent Thereby we provide a comprehensive survey of recent advances in NN-based active text classification. 
Having reviewed these recent advances, we illuminate areas that either need re-evaluation, or have not yet been evaluated in a more recent context.
As a final result, we develop research questions outlining the scope of future research.

%

%% file: tex/related_work.tex
\section{Related Work}

\textcite{settles2010active} provides a general active learning survey, 
summarizing the prevalent AL scenario types and query strategies.
%
They present variations of the basic AL setup like 
 variable labeling costs or alternative query types, 
and most notably, they discuss empirical and theoretical research investigating the effectiveness of AL: 
They mention research suggesting that AL is effective in practice 
and has increasingly gained adoption in real world applications. However, it is pointed out that empirical research also reported cases in which AL performed worse than passive learning and that the theoretical analysis of AL is incomplete.
Finally, relations to related research areas are illustrated, thereby connecting AL among others to reinforcement learning and semi-supervised learning.

\hiddennotes{
  settles2009/settles2010active \cite{settles2010active}

  \begin{itemize}
    \item scenarios: membership query synthesis, stream-based, pool-based
    \item query strategies
    \begin{itemize}
      \item keine Klassifikation vorgenommen
    \end{itemize}
    \item most important (at that time): analysis
    \begin{itemize}
      \item does it work?
      \item both empirical and theoretical evidence
    \end{itemize}
    \item TODO: problem setting variants, practical considerations (only in 2010 version)
    \item related areas: semi-supervised, reinforcement, submodular optimization, equivalence query learning, model parroting and compression
  \end{itemize} 
}

The survey of \textcite{fu2013survey} is focused around a thorough analysis of uncertainty-based query strategies, which are categorized into a taxonomy.
This taxonomy differentiates at the topmost level between the uncertainty of i.i.d. instances and instance correlation. The latter is a superset of the former 
and intends to reduce redundancy among instances by considering feature, label, and structure correlation when querying.
Moreover, they perform an algorithmic analysis for each query strategy 
and order the strategies by their respective time complexity, highlighting the increased complexity for correlation-based strategies.

\hiddennotes{
  \begin{itemize}
    \item survey on instance selection
  \end{itemize} 
}

Another general survey covering a wide range of topics was conducted by \textcite{aggarwal2014active}. 
They provide a flat categorization of query strategies, which is quite different from the taxonomy of \textcite{fu2013survey} and divides them into the following three categories: (1) ``heterogenity-based'', which sample instances by their prediction uncertainty or dissimilarity compared to existing labeled instances, (2) ``performance-based''', which select instances based on a predicted change of the model loss, and (3) ``representativeness-based'', which select data points to reflect a larger set in terms of their properties, usually achieved by the means of distribution density \cite{aggarwal2014active}.
Similarly to \cite{settles2010active}, they present and discuss many non-standard variations of the active learning scenario.

\revised{
  An NLP-focused active learning survey was performed by \textcite{olsson2009literature}.
  This work's main contribution is a survey of disagreement-based query strategies, which use the disagreement among multiple classifiers to select instances.
  Moreover, Olsson reviews practical considerations, e.g., selecting an initial seed set,
  deciding between stream-based and pool-based scenario, and deciding when to terminate the learning process.
  %

  %
  %
  Although some NN-based applications are mentioned, none of the above surveys covers NN-based AL in depth.
  Besides, none is recent enough to cover NN-architectures, which have only recently been adapted successfully to text classification problems like e.g., KimCNN \cite{kim2014convolutional}.
  The same holds true for recent advances in NLP such as word embeddings, contextualized language models (explained in Section \ref{subsect:tc_recent_adv}), or resulting advances in text classification (discussed in Section \ref{subsect:nn_text_classification} and Section \ref{subsect:active_tc}).
  We intend to fill these gaps in the remainder of this survey.
}

\hiddennotes{
  \begin{itemize}
    \item NLP-centric Active Learning Survey
    \item describes multiple active learning modes: Query by uncertainty, query by committee, redundant views
    \item extensively reviews strategies to measure disagrement
  \end{itemize}
}

%% file: tex/active_learning.tex
\section{Active Learning}
\label{sect:active_learning}

\begin{figure}[!h]
  \centering 
  \includegraphics{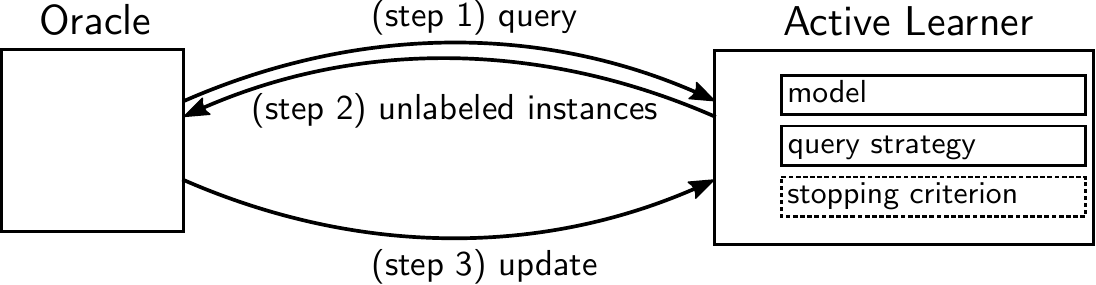}
  \caption{An overview of the AL process: Model, query strategy, and (optionally a) stopping criterion are the key components of an active learner.
  The main loop is as follows:
  First the oracle queries the active learner, which returns a fixed amount of unlabeled instances.
  Then, for all selected unlabeled unstances are assigned labels by the oracle.
  This process is repeated until the oracle stops, or a predefined stopping criterion is met.}
  \label{fig:al_process}
\end{figure}

The goal of AL is to create a model using as few labeled instances as possible,
i.e. minimizing the interactions between the oracle and the active learner.
The AL process (illustrated in Figure \ref{fig:al_process}) is as follows: The oracle requests unlabeled instances from the active learner ({\em query}, see Figure \ref{fig:al_process}: step 1), 
which are then selected by the active learner (based on the selected query strategy) and passed to the oracle (see Figure \ref{fig:al_process}: step 2).
Subsequently, these instances are labeled by the oracle and returned to the active learner ({\em update}, see Figure \ref{fig:al_process}: step 3).
After each update step the active learner's model is retrained, which makes this operation at least as expensive as a training of the underlying model.
This process is repeated until a stopping criterion is met (e.g., a maximum number of iterations or a minimum threshold of change in classification accuracy).

\noindent The most important component for AL is the query strategy.
In the introduction we claimed that a large fraction of query strategies are uncertainty-based.
To analyze this we provide a taxonomy of query strategies in the following section and highlight the parts in which uncertainty is involved.
For a general and more detailed introduction on AL refer to the surveys of \textcite{settles2010active} and \textcite{aggarwal2014active}.

\subsection{Query Strategies}
\label{subsec:query_strategies}

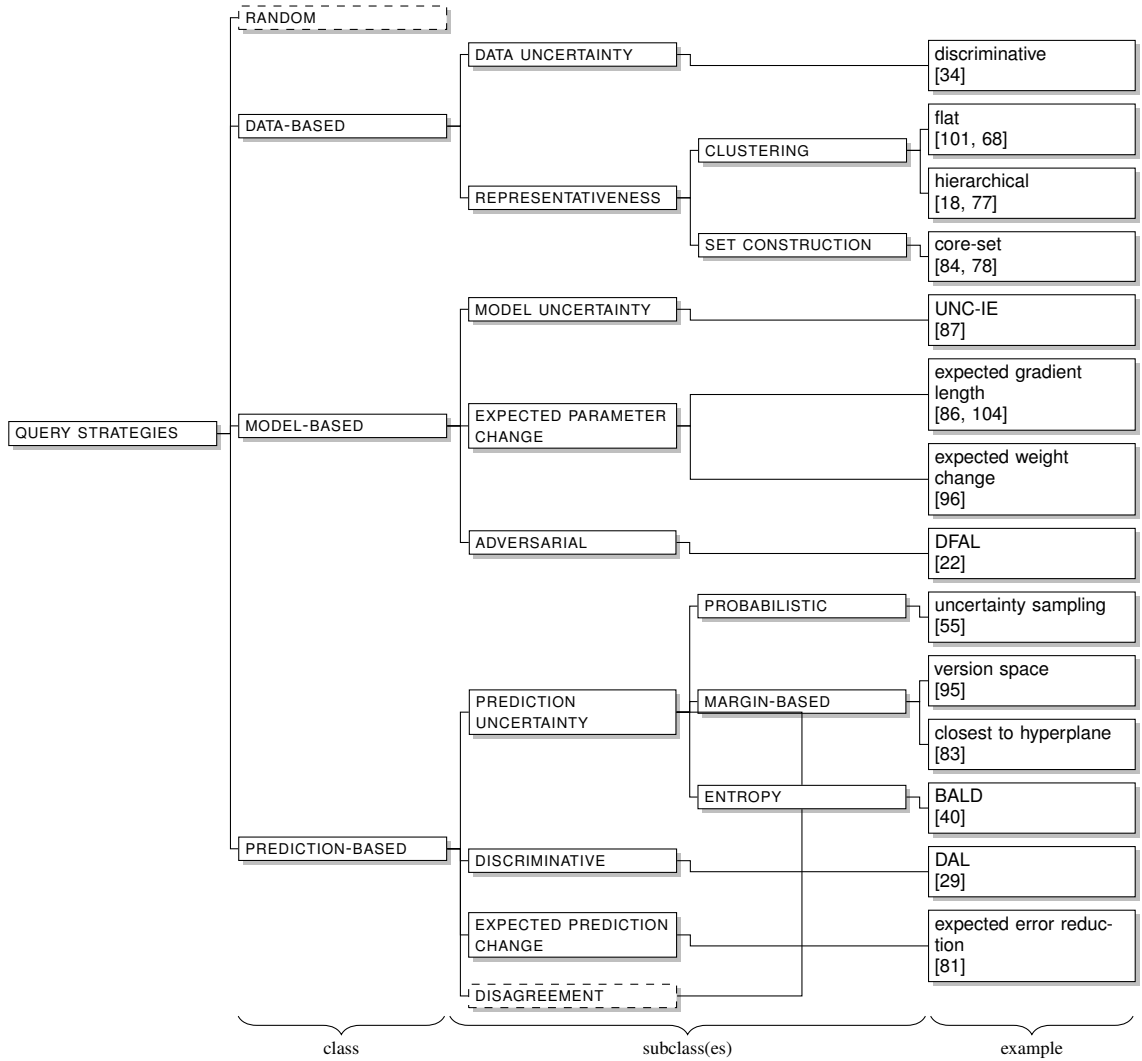
\begin{figure}[!h]
  {
    \scriptsize
    \centering
    \begin{forest} for tree={
        grow'=0, draw
      },
      forked edges,
      [\scshape query strategies, no edge, root
          [\scshape random, onode_dashed
              [,phantom]
          ]
          [\scshape data-based, onode
              [\scshape data uncertainty, onode
                [discriminative\\ \cite{guo2008discriminative}, tier=leaf, tnode]
              ]
              [\scshape representativeness, onode
                [\scshape clustering, onode
                    [flat\\ \cite{xu2003representative,nguyen2004active}, tier=leaf, tnode]
                    [hierarchical\\ \cite{dasgupta2008hierarchical,poursabzi-sangdeh2016alto}, tier=leaf, tnode]
                ]
                [\scshape set construction, onode
                  [core-set\\ \cite{sener2017active,prabhu2019sampling}, tier=leaf, tnode]
                ]
              ]
          ]
          [\scshape model-based, onode
              [\scshape model uncertainty, onode
                  [UNC-IE\\ \cite{sharma2017evidence-based}, tier=leaf, tnode]
              ]
              [\scshape expected parameter\\change, onode
                  [expected gradient length\\ \cite{settles2008multiple-instance,zhang2017active}, tier=leaf, tnode]
                  [expected weight change\\ \cite{vezhnevets2012active}, tier=leaf, tnode]
              ]
              [\scshape adversarial, tnode
                  [DFAL\\ \cite{ducoffe2018adversarial}, tier=leaf, tnode]
              ]  
          ]
          [\scshape prediction-based, name=prediction_based, onode
              [\scshape prediction\newline uncertainty, name=prediction_uncertainty, tnode
                  [\scshape probabilistic, onode
                      [uncertainty sampling \cite{lewis1994sequential}, tier=leaf,tnode]
                  ]
                  [\scshape margin-based, onode
                      [version space\\ \cite{tong2001support}, tier=leaf,tnode]
                      [closest to hyperplane\\ \cite{schohn2000less}, tier=leaf,tnode]
                  ]
                  [\scshape entropy, name=entropy, onode
                      [BALD\\ \cite{houlsby2011bayesian}, tier=leaf,tnode]
                  ]
              ]
              [\scshape discriminative, onode
                  [DAL\\ \cite{gissin2019discriminative}, tier=leaf , tnode]
              ]           
              [\scshape expected prediction change\\, onode
                  [expected error reduction\\ \cite{roy2001toward}, name=lastnode, tier=leaf, tnode]
              ]
              [\scshape disagreement, name=disagreement, onode_dashed
                  [,phantom, tier=leaf]
              ]
          ] 
      ]
      \draw[zlevel,-] (disagreement.east) -| (entropy.south);
      \draw[zlevel,-] (entropy.north) |- (prediction_uncertainty.east);
      \path (current bounding box.south) coordinate (s1);
      \draw[decorate,decoration={brace,amplitude=1em,mirror}] ([yshift=-3pt]prediction_based.south west|-s1) -- node[below=1em] {class} ([yshift=-3pt,xshift=-1pt]prediction_based.south east|-s1);
      \draw[decorate,decoration={brace,amplitude=1em,mirror}] ([yshift=-3pt,xshift=1pt]prediction_based.south east|-s1) -- node[below=1em] {subclass(es)} ([yshift=-3pt,xshift=-1pt]lastnode.south west|-s1);
      \draw[decorate,decoration={brace,amplitude=1em,mirror}] ([yshift=-3pt,xshift=1pt]lastnode.south west|-s1) -- node[below=1em] {example} ([yshift=-3pt,xshift=-1pt]lastnode.south east|-s1);
    \end{forest}
     
    \caption{\revised{A taxonomy of query strategies for AL.
      The key distinction is at the first level, where the query strategies are categorized by their access to different kinds of input information. 
      From the second to the penultimate level we form coherent subclasses,
      and the final level shows examples for the respective class.
      This taxonomy is not exhaustive due to the abundance of existing query strategies, and it is biased towards query strategies in NLP.}
      }
    \label{fig:query_strategies}
  }
\end{figure}

In Figure \ref{fig:query_strategies} we classify the most common AL query strategies based on a strategy's {\em input information}, which denotes the numeric value(s) a strategy operates on.
In our taxonomy the input information can be either random or one of data, model, and prediction.
\phrasing{These categories are ordered by increasing complexity and are not mutually exclusive.}
\phrasing{Obviously, the model is a function of the data, as well as the prediction is a function of model and data, and moreover, in many cases a strategy use multiple of these criteria.}
In such cases we assign the query strategy to the most specific category (i.e. prediction-based precedes model-based, which in turn precedes data-based).

\hiddennotes{
  \begin{itemize}
    \item Table in appendix: For each strategy here: Applicability: B / MC / ML\\
  \end{itemize}
}

\paragraph{Random}

Randomness has traditionally been used as a baseline for many tasks.
In this case, random sampling selects instances at random and is a strong baseline for AL instance selection \cite{lewis1994sequential,schohn2000less,roy2001toward}.
\revised{It often performs competitive to more sophisticated strategies, especially when the labeled pool has grown larger \cite{sener2017active,ducoffe2018adversarial}.}
\paragraph{Data-based} Data-based strategies have the lowest level of knowledge, i.e. they only operate on the raw input data and optionally the labels of the labeled pool. 
We categorize them further into (1) strategies relying on data-uncertainty, which may use information about the data distribution, label distribution, and label correlation,
and (2) representativeness, \revised{which tries to geometrically compress a set of points, by using fewer representative instances to represent the properties of the entirety}.
%

%
%

\paragraph{Model-based} 
The class of model-based strategies has knowledge about both the data and the model.
These strategies query instances based on measure provided by the model given an instance. 
An example for this would be a measure of confidence for the model's explanation of the given instance \cite{gal2016uncertainty},
for example, how reliable the model rates encountered features.
This can also be an expected quantity, for example in terms of the gradient's magnitude \cite{settles2008multiple-instance}. 
While predictions from the model can still be obtained, we impose the restriction that the target metric must be an (observed or expected) quantity of the model, excluding the final prediction.
\revised{Model-based uncertainty is a noteworthy subclass here, which operates using the uncertainty of a model's weights \cite{gal2016uncertainty}.}
\revised{\textcite{sharma2017evidence-based} describe a similar class, in which the uncertainty stems from not finding enough evidence in the training data, i.e. failing to separate classes at training time.}
\revised{They refer to this kind of uncertainty as {\em insufficient evidence uncertainty}.}

\paragraph{Prediction-based}
Prediction-based strategies select instances by scoring their prediction output.
The most prominent members of this class are prediction-uncertainty-based and disagreement-based approaches.
\textcite{sharma2017evidence-based} denote prediction-based uncertainty by {\em conflicting-evidence uncertainty}, which they, contrary to this work, count as another form of model-based uncertainty.
There is sometimes only a thin line between the concepts of model-based und prediction-based uncertainty.
\revised{Roughly speaking, prediction-based uncertainty corresponds in a classification setting to inter-class uncertainty, as opposed to model-based uncertainty, which corresponds to intra-class uncertainty.}
In literature, uncertainty sampling \cite{lewis1994sequential} usually refers to prediction-based uncertainty, unless otherwise specified.
\paragraph{Ensembles} \revised{When a query strategy combines the output of multiple other strategies, this is called an {\em ensemble}.}
\revised{We only classify the concept of ensemble strategies within the taxonomy (see disagreement-based subclass in Figure \ref{fig:query_strategies}) without going into detail due to several reasons:}
(1) Ensembles are again composed of primitive query strategies, which can be classified using our taxonomy.
(2) Ensembles can be hybrids, i.e. they can be a mixture of different classes of query strategies.
Moreover, the output of an ensemble is usually a function of the disagreement among the single classifiers, which is already covered in previous surveys of \textcite{olsson2009literature} and \textcite{fu2013survey}.\\

\noindent We are not the first to provide a classification of query strategies: \textcite{aggarwal2014active} provide an alternative classification, which divides the query strategies into heterogenity-based models, 
performance-based models, and representativeness-based models.
Heterogenity-based models try to sample diverse data points, w.r.t the current labeled pool. This class includes among others uncertainty sampling and ensembles, i.e. no distinction is made between ensembles and single-model strategies. 
Performance-based models aim to sample data targeting an increase of the models performance, for example a reduction in the model's error.  
This intersects with our model-based class, however, it lacks strategies which focus on a change of parameters (e.g., expected gradient length \cite{settles2008multiple-instance}) as opposed to changes in a metric.
Lastly, representativeness-based strategies sample instances so that the distribution of the subsample is as similar as possible to the training set.
Although similar to our data-based class, they always assume the existence of a model, which is not the case for data-based strategies.\\
\pindent \textcite{fu2013survey} separate query strategies into 
uncertainty-based and diversity-based classes. 
Uncertainty-based strategies assume the i.i.d. distribution of instances; they compute a separate score for each instance, which is the basis for the instance selection.
Diversity-based strategies are a superset thereof and additionally consider correlation amongst instances.
Thereby they characterize uncertainty and correlation as critical components for query strategies.
This classification successfully distinguishes query strategies by considering exclusively uncertainty and correlation.
However, it is less transparent in terms of the input information, 
which our taxonomy highlights.
Nevertheless, correlation is a factor orthogonal to our taxonomy and can be added as an additional criterion.
\\
\pindent After creating our taxonomy, we discovered a recent categorization of uncertainty in deep learning \cite{gal2016uncertainty}, which distinguishes between data-, model-, and predict{\em{}ive}-{\em uncertainty}, similar to the taxonomy's first level (data-, model-, prediction-based query strategies).
Although this classification comes naturally from the data's degree of processing, we emphasize that we are not the first to come up with this abstraction.

By using the input information as decisive criterion, this taxonomy provides an information-oriented view on query strategies.
It highlights in which parts and how uncertainty has been involed in existing query strategies. Uncertainty in terms of NNs is, however, is known to be challenging as described in Section \ref{subsect:nn_al}. Moreover, we use the taxonomy to categorize recent work in AL for text classification in Section \ref{subsec:experiments}.




\subsection{Neural-Network-Based Active Learning}
\label{subsect:nn_al}

In this section we investigate the question, why neural networks are not more prevalent in AL applications. 
This can be attributed to two central topics: Uncertainty estimation in NNs, and the contrast of NNs requiring between big data and AL dealing with small data.
We examine these issues from a NN perspective, alleviating the NLP focus.

\paragraph{Previous Work} 

\hiddennotes{
  \begin{itemize}
    \item Überleitung fehlt
    \item Argumentationsstruktur unklar?
    \begin{itemize}
      \item Classic -> RNN -> CNN -> LSTM -> GANs (-> Meta?) 
    \end{itemize}
  \end{itemize}
}

Early research in NN-based AL can be divided into uncertainty-based \cite{cohn1994improving},
and ensemble-based \cite{krogh1995neural,melville2004diverse} strategies.
The former often use prediction entropy \cite{mackay1992evidence,roy2001toward} as measure of uncertainty,
while the latter utilize the disagreement among the single classifiers.
%
%
%
%
\textcite{settles2008multiple-instance} proposed the expected gradient length (EGL) query strategy, which selects instances by the  expected change in the model's weights. \textcite{zhang2017active} were first to use a CNN for AL.
They proposed a variant of the expected gradient length strategy \cite{settles2008multiple-instance}, in which they select instances that are expected to result in the largest change in embedding space, thereby training highly discriminative representations.
\textcite{sener2017active} observed uncertainty-based query strategies not to be effective for CNN-based batch-mode AL, and proposed core-set selection, which samples a small subset to represent the full dataset. \textcite{ash2019deep} proposed BADGE, a query strategy for DNNs, which uses k-means++ seeding \cite{arthur2007k} on the gradients of the final layer, in order to query by uncertainty and diversity.

%
%

Finally, Generative Adversarial Networks (GANs; \cite{goodfellow2014generative}) have also been applied successfully for AL tasks: 
\textcite{zhu2017generative} use GANs for query synthesis of images within an active learner using an SVM model.
The instances are synthesized so that they would be classified with high uncertainty. 
The authors report this approach to outperform random sampling, pool-based uncertainty sampling using an SVM \cite{tong2001support}, and in some cases passive learning, while having the weakness to generate too similar instances.
The approach itself is neither pure NN-based, nor does it belong to the pool-based scenario, however, it is the first reported use of GANs for AL.
\textcite{ducoffe2018adversarial} use adversarial attacks to find instances that cross the decision boundary with the aim to increase the model robustness.
They train two CNN architectures and report results superior to the core-set \cite{sener2017active} strategy on image classification tasks.
It is obvious that GANs inherently belong to the membership query synthesis scenario.
\revised{Therefore their performance correlates with the quality of artificial data synthesis, i.e. they are usually not that effective for NLP tasks. This has already been recognized and first improvements towards a better text generation have been made \cite{zhang2017adversarial}}.
%

%
%

%
\paragraph{Uncertainty in Neural Networks} One of the earliest and in many variations adopted class of strategies is uncertainty sampling \cite{schohn2000less,tong2001support}.
\label{sect:nn_uncertainty}
Unfortunately, this widely-used concept is not straightforward to apply for NNs, as they do not provide an inherent indicator of uncertainty.
In the past, this has been tackled among others by ensembling \cite{krogh1995neural,heskes1996practical,carney1999confidence}, or by learning error estimates \cite{nix1995learning}.
More recent approaches furthermore use Bayesian extensions \cite{blundell2015weight}, 
obtain uncertainty estimations using dropout \cite{srivastava2014dropout,gal2016dropout},
or use probabilistic NNs to estimate predictive uncertainty \cite{lakshminarayanan2017simple}.
However, ensemble and Bayesian approaches quickly become infeasible on larger datasets,
and NN architectures are generally known to be overconfident in their predictions \cite{guo2017calibration,lakshminarayanan2017simple}.
Consequently, uncertainty in NNs is only insufficiently solved and therefore still remains a highly relevant research area.

\hiddennotes{
  To circumvent this, various approaches have been suggested:
  (1) Bayes by Backprop \cite{blundell2015weight} uses Bayesian variational inference in order to learn approximated distributions over the weights of a NN.
  (2)  extend dropout \cite{srivastava2014dropout} to represent uncertainties 
  by interpreting dropout-enhanced NNs as a Bayesian approximation to deep Gaussian Processes.
  However, this has been reported to not scale well for larger datasets \cite{sener2017active}.
  (3) Traditionally this has been tackled by ensembling \cite{krogh1995neural}, i.e. producing $k$-many different models and outputting a prediction by interpreting their disagreement (see \textcite{olsson2009literature}).
  %
  %
  The disadvantage of this approach is the obvious additional cost in runtime and model size.
  %
  %
  \textcite{sener2017active} report uncertainty-based approaches to be not effective for CNNs, and \textcite{ducoffe2018adversarial} observed worse performance compared to random sampling.\\
}

%
%
\paragraph{Contrasting Paradigms} 
DNNs are known to excel in particularly at large-scale datasets,
but often having large amounts of data available is a strict requirement to perform well at all (e.g., \cite{zhang2016character-level}).
AL on the other hand tries to minimize the labeled data.
The small labeled datasets can be a problem for DNNs, since they are known to overfit on small datasets (e.g., \cite{taylor2017improving,wang2017effectiveness}), 
which results in bad generalization performance on the test set.
Moreover, DNNs often offer little advantage over shallow models when they are trained using small datasets \cite{shen2018deep}, thereby lacking justification for their higher computational costs.
On the other hand we clearly cannot require AL to label more data, since this would defeat its purpose.
Therefore there has been research on dealing with (D)NNs using small datasets, however, it is only a scarce amount, especially in relation to the large volume of NN literature in general.
Handling small datasets is mostly circumvented by using pre-training \cite{hinton2006reducing,wagner2013learning} or other transfer learning approaches \cite{caruana1995learning,bengio2012deep,wagner2013learning}.
%
Finally, the search for optimal hyperparameters is often neglected and instead the hyperparameters of related work are used, which are optimized for large datasets, if at all.

%% file: tex/text_classification.tex
\section{Active Learning for Text Classification}
\label{sect:text_classification}

In Sections \ref{subsect:tc_recent_adv} and \ref{subsect:active_tc} we first summarize recent methods in text classification and NNs.
We elaborate on each method's importance in the context of AL, and analyze its adoption by recent research where applicable.
For insufficiently adopted methods, we present how they could advance AL for text classification.
Most importantly, we present an overview of recent experiments in AL for text classification and analyze commonalities and shortcomings. 

\subsection{Recent Advances in Text Classification}
\label{subsect:tc_recent_adv}
%
%
\paragraph{Representations} Traditional methods use the bag-of-words (BoW) representation, which are sparse and high-dimensional.
However, with the introduction of word embeddings like word2vec \cite{mikolov2013efficient,mikolov2014distributed}, 
GloVe \cite{pennington2014glove}, or fastText \cite{joulin2016bag},
word embeddings have replaced BoW representations in many cases.
This is due to several reasons:
(1) They represent semantic relations in vectors space and avoid the problem of mismatching features as for example due to synonymy;
(2) incorporating word embeddings resulted in superior performance for many downstream tasks \cite{mikolov2013efficient,pennington2014glove,joulin2016bag};
(3) unlike bag-of-words, word vectors are dense, low-dimensional representations, 
which makes them applicable to a wider range of algorithms --
especially in the context of NNs which favor fixed-size inputs.
Various approaches have been presented in order to obtain similar fixed size representations for word sequences, i.e. sentences, paragraphs or documents \cite{le2014distributed}.\\
\pindent Word embeddings are representations, which provide exactly one vector per word and in consequence one meaning as well.
This makes them also unaware of the current word's context and therefore makes them unable to detect and handle ambiguities.
Unlike word embeddings, language models (LMs) compute the word vector using the word and the surrounding context \cite{peters2018deep}.
This results in a contextualized representation, which inherits the advantages of word embeddings, and at the same time allows for context-specific representation (in contrast to static embeddings) \cite{peters2018deep}. 
ELMo was the first LM to gain wide adoption and surpassed state of the art models on several NLP tasks \cite{peters2018deep}.
Shortly thereafter, BERT \cite{devlin2019bert} was introduced and provided bidirectional pre-training-based language modelling.
The process to create a BERT-based model consists of a pre-training and a fine-tuning step as opposed to ELMO's direct feature-based approach in which contextualized vectors are obtained from the pre-trained model and used directly as features \cite{devlin2019bert}.
By masking, i.e. randomly removing a fraction of tokens during training, the training was adapted to predict the masked words.
This made the bidrectional training possible, which would otherwise be obstructed because a word could "see itself" when computing its probability of occurrence given a context \cite{devlin2019bert}.
Following this, XLNet \cite{yang2019xlnet} introduced a similar approach of pre-training and fine-tuning using an autoregressive language model, however, it overcame BERT's limitation as it does not rely on masking data during pre-training \cite{yang2019xlnet}, and moreover, successfully manages to integrate the recent TransformerXL architecture \cite{dai2019transformer}.
Since then, a variety of LMs have been published, which further optimize the pre-training of previous LM architectures (e.g., RoBERTa \cite{liu2019roberta} and ELECTRA \cite{clark2020electra}), or distill the knowledge into a smaller model (e.g., DistilBERT \cite{sanh2020distilbert}).
Similarly to word embeddings, there are approaches to use LMs in order to obtain sentence representations from LMs \cite{reimers2019sentence-bert}.\\
\pindent All mentioned representations offer a richer expressiveness than traditional BoW representations and therefore are well-suited for active learning purposes.


\paragraph{Neural-Network-Based Text Classification} 
\label{subsect:nn_text_classification}

A well-known CNN architecture presented by \textcite{kim2014convolutional} (KimCNN) operates on pre-trained word vectors and achieved state of the art results at the time using only a simple but elegant architecture.
The investigated CNN setups did not require much hyperparameter tuning and confirmed the effectiveness of dropout \cite{srivastava2014dropout} as a regularizer for CNN-based text classification.\\
%
%
\pindent The word embeddings of fastText \cite{joulin2016bag} differ from other word embeddings in the sense that the approach is (1) supervised and (2) specifically designed for text classification.
Being a shallow neural network, it is still very efficient, while still obtaining performances comparable to deep learning approaches at that time.\\
%
%
%
\pindent\textcite{howard2018universal} developed Universal Language Model Fine-tuning (ULMFiT), a LM transfer learning method using the AWD-LSTM architecture \cite{merity2017regularizing},
which outperformed the state of the art on several text classification datasets when trained on only $100$ labeled examples, and thereby achieved results significantly superior to more sophisticated architectures of previous work.
%
%
%
%
%
%
Context-specific LMs like BERT \cite{devlin2019bert} and XLNet \cite{yang2019xlnet} yield a context-dependent vector for each token, thereby strongly improving NN-based text classification \cite{devlin2019bert,yang2019xlnet,sun2020how}.
%
%
State of the art in NN-based text classification is LM-based fine-tuning with XLNet, which has a slight edge over BERT in terms of test error rate \cite{yang2019xlnet,sun2020how}.
ULMFiT follows closely thereafter, and KimCNN is still a strong contender.
Notably, ULMFiT, BERT and XLNet all perform {\em transfer learning}, which aims to transfer knowledge from one model to another \cite{pratt1991direct,caruana1995learning},
thereby massively reducing the required amounts of data.
%
%

%
%

\subsection{Text Classification for Active Learning}
\label{subsect:active_tc} 

Traditional AL for text classification heavily relied on query strategies based on prediction-uncertainty \cite{lewis1994sequential} and ensembling \cite{liere1997active}. 
Common model choices included support vector machines (SVMs; \cite{tong2001support}), naive bayes \cite{nigam2000text}, logistic regression \cite{hoi2006large-scale} and neural networks \cite{krogh1995neural}.
To the best of our knowledge, no previous survey covered traditional AL for text classification, however,  ensembling-based AL for NLP has been covered in depth by \textcite{olsson2009literature}.\\
\pindent Regarding modern NN-based AL for text classification, the relevant models are primarily CNN- and LSTM-based deep architectures: \textcite{zhang2017active} claim to be the first to consider AL for text classification using DNNs.
They use CNNs and contribute a query strategy, which selects the instances \phrasing{based on the expected change of the word embeddings and the model's uncertainty given the instance,} thereby learning discriminative embeddings for text classification.
\textcite{an2018deep} evaluated SVM, LSTM and gated recurrent unit (GRU; \textcite{cho2014properties}) models, and reported that the latter two significantly outperformed the SVM baseline on the Chinese news dataset ThucNews.
\textcite{lu2020investigating} investigated the performance of different text representations in a pool-based AL scenario.
They compared frequency-based text representations, word embeddings and transformer-based representations used as input features for a SVM-based AL and different query strategies,
in which transformer-based representations yielded consistently higher scores.
\textcite{prabhu2019sampling} investigate sampling bias and apply active text classification on the large scale text corpora of \textcite{zhang2016character-level}.
They demonstrate FastText.zip \cite{joulin2016fasttextzip} with (entropy-based) uncertainty sampling to be a strong baseline, which is competitive compared to recent approaches in active text classification.
Moreover, they use this strategy to obtain a surrogate dataset (comprising from 5\% to 40\% of the total data) on which a LSTM-based LM is trained using ULMFiT \cite{howard2018universal}, reaching accuracy levels close to a training on the full dataset.
Unlike past publications, they report this uncertainty-based strategy to be effective, robust, and at the same time computationally cheap.
This is the most relevant work in terms of \revised{the intersection between text classification, NNs and DL}.


%
%

\hiddennotes{
    \begin{itemize}
        \item Previous Research nach Query Strategy Class / Embeddings untersuchen
        \item Reinforcement Learning using NNs
    \end{itemize}
}

%
%
\begin{table}[h!]
  \begin{center}
  \begin{tabular}{l P{3cm} P{2.5cm} P{4.5cm}}
    \hline
    \textbf{Publication} & \textbf{Datasets} & \textbf{Model(s)} & \textbf{Query Strategy Class(es)}\\
    \hline
    \mbox{\cite{hu2016active}} & 20N, R21, RV2, SPM & NB, SVM, kNN & 1. Prediction uncertainty (LC)\newline 2. Prediction uncertainty (CTH)\newline 3. Prediction uncertainty (disagreement) \\
    \mbox{\cite{zhang2017active}} & CR, MR, SJ, MRL, MUR, DR & CNN & 1. Model uncertainty (EGL)\newline 2. Prediction Uncertainty (entropy)\\
    \mbox{\cite{bloodgood2018support}} & RMA & SVM & 1. Prediction uncertainty (CTH)\newline 2. Prediction uncertainty (disagreement)\\
    \mbox{\cite{siddhant2018deep}} & TQA, MR & SVM, CNN, BiLSTM & Prediction uncertainty (disagreement)\\
    \mbox{\cite{lowell2019practical}} & MR, SJ, TQA, CR & SVM, CNN, BiLSTM & 1. Prediction uncertainty (entropy)\newline 2. Prediction uncertainty (disagreement)\\
    \mbox{\cite{prabhu2019sampling}} & SGN, DBP, YHA, YRP, YRF, AGN, ARP, ARF & FTZ, ULMFiT & Prediction uncertainty (entropy)\\
    \mbox{\cite{lu2020investigating}} & MRL, MDS, BAG, G13, ACR, SJ, AGN, DBP &  SVM & 1. Prediction uncertainty (CTH) \newline 2. Prediction uncertainty (disagreement) \newline 3. Data-based (EGAL) \newline 4. Data-based (density)\\
    \hline 
  \end{tabular}
  \captionsetup{aboveskip=10pt}
  \caption{An overview of recent work on AL for text classification.
    We referred to the datasets using short keys, which can be looked up in Table \ref{tab:datasets} in the Appendix. Models: Naive Bayes (NB), Support Vector Machine (SVM), k-Nearest Neighbours (kNN), Convolutional Neural Network (CNN), [Bidirectional] Long Short-Term Memory ([Bi]LSTM), FastText.zip (FTZ), Univeral Language Model Fine-Tuning (ULMFiT). Query strategies: Least confidence (LC), Closest-to-hyperplane (CTH), expected gradient length (EGL).
    Random selection baselines were omitted.}
  \label{table:recent_work}
\end{center}
\end{table}

\subsection{Commonalities and Limitations of Previous Experiments}
\label{subsec:experiments}

Table \ref{table:recent_work} shows the most recent AL for text classification experiments, all of them more recent than the surveys of \textcite{settles2010active} and \textcite{olsson2009literature}. 
For each publication we list the utilized datasets, models, and classes of query strategies (with respect to the taxonomy in Section \ref{subsec:query_strategies}).
%
%
We present this table in order to get insights about the recently preferred classification models and query strategy classes.\\
\pindent We can draw multiple conclusions from Table \ref{table:recent_work}:
It is obvious that a significant majority of these query strategies belong to the class of prediction-based query strategies, more specifically to the prediction-uncertainty and disagreement-based sub-classes. 
In addition to that, we can identify several shortcomings:
First, in many experiments two or more standard datasets are evaluated, but very often there is little to no intersection between the experiments in terms of their datasets. 
As a result we lose comparability against previous research.
For recent research, this can seen in Table \ref{table:recent_work}, where the only larger intersections is between the works of \textcite{zhang2017active} and \textcite{lowell2019practical}.
\textcite{siddhant2018deep} provide at least some comparability against \textcite{zhang2017active} and \textcite{lowell2019practical} through one dataset each.
Additionally, RMA \cite{apte1994towards} is a subset of R21 \cite{lewis1997reuters}, which are used by \textcite{bloodgood2018support} and \textcite{hu2016active}, so they might be comparable to some degree.
\cite{prabhu2019sampling} are the only ones to evaluate on the more recent large-scale text classification datasets \cite{zhang2016character-level}, and although these datasets are more realistic in terms of their size, the authors omitted the classic datasets, so it is difficult to relate their contributions to  previous work.
Moreover, as a result of this, we do not know if and to what degree past experiments generalize to DNNs \cite{prabhu2019sampling}.

\noindent Finally, it is not clear if recent (D)NNs benefit from the same query strategies, i.e. past findings may not apply to modern NN architectures:
\textcite{prabhu2019sampling} identified contradicting statements in recent literature about the effectiveness of using prediction uncertainty in combination with NNs.
They achieved competitive results using a FastText.zip (FTZ) model and a prediction uncertainty query strategy, which proved to be very effective while requiring only a small amount of data, despite all reported weaknesses concering NNs and uncertainty estimates.

%% file: tex/open_research_questions.tex
\section{Open Research Questions}

\paragraph{Uncertainty Estimates in Neural Networks}

In Section \ref{sect:active_learning} it was illustrated that uncertainty-based strategies 
have been used successfully in combination with non-NN models, and in Section \ref{subsec:experiments} it was shown that they also account for the largest fraction of query strategies in recent NN-based AL. 
Unfortunately, uncertainty in NNs is still challenging due to inaccurate uncertainty estimates, or limited scalability (as described in Section \ref{sect:nn_uncertainty}). 

\paragraph{Representations}

As outlined in Section \ref{subsect:tc_recent_adv}, the use of text representations in NLP has shifted from bag-of-words to static and contextualized word embeddings.
These representations evidentially provide many advantages like disambiguation capabilities, non-sparse vectors, and an increase in performance for many tasks.
Although there have been some applications \cite{zhang2017active,prabhu2019sampling,lu2020investigating}, there is no AL-specific systematic evaluation to compare word embeddings and LMs using NNs.
Moreover, they are currently only scarcely used,
which hints at either a slow adoption, or some non-investigated practical issues.

\paragraph{Small Data DNNs}

DL approaches are usually applied in the context of large datasets.
AL, however, necessarily intends to keep the (labeled) dataset as small as possible.
In Section \ref{sect:active_learning} we outlined why small datasets can be challenging for DNNs, and as a direct consequence as well for DNN-based AL.
Using pre-trained language models, this problem is alleviated to some degree because fine-tuning allows training models using considerably smaller datasets.
%
Nonetheless, it is to be investigated how little data is still necessary to successfully fine-tune a model.

\paragraph{Comparable Evaluations}

In Section \ref{subsec:experiments} we provided an overview of the most common AL strategies for text classification.
Unfortunately, the combinations of datasets used in the experiments are often completely disjoint, e.g. \textcite{siddhant2018deep}, \textcite{lowell2019practical}, and \textcite{prabhu2019sampling}.
As a consequence, comparability is decreased or even lost, especially between more recent and past work.
Comparibility is, however, crucial to verify if past insights regarding shallow NN-based AL still apply in context of DNN-based AL \cite{prabhu2019sampling}.

\paragraph{Learning to Learn}

There is an abundance of query strategies to choose from, which we have (non-exhaustively) categorized in Section \ref{subsec:query_strategies}.
This introduces the problem of choosing the optimal strategy.
The right choice depends on many factors like data, model, or task, and can even vary between different iterations during the AL process.
As a result, {\em learning to learn} (or {\em meta-learning}) has become popular and can be used to learn the optimal selection \cite{hsu2015active}, or even learn query strategies as a whole \cite{bachman2017learning,konyushkova2017learning}.

%% file: tex/conclusions.tex
\section{Conclusions}

In this survey, we investigated (D)NN-based AL for text classification and inspected factors obstructing its adoption.
We created a taxonomy, distinguishing query strategies by their reliance on data-based, model-based, and prediction-based input information.
We analyzed query strategies used in AL for text classification and categorized them into the respective taxonomy classes.
We presented the intersection between AL, text classification and DNNs, which is to the best of our knowledge the first survey of this topic.
Furthermore, we reviewed (D)NN-based AL, identified current challenges and state of the art, and pointed out that it is both underresearched and often lacks comparability.
In addition to that, we presented relevant recent advances in NLP, related them to AL, and showed gaps and limitations for their application.
One of our main findings is that uncertainty-based query strategies are still the most widely used class, regardless of whether the analysis is restricted to NNs.
LM-based representations offer finer-grained context-specific representations while also handling out-of-vocabulary words. 
Moreover, we find fine-tuning-based transfer learning alleviates the small data problem to some degree but lacks adoption.
Most important DNNs are known for their strong performance on many tasks and first adoptions in AL have shown promising results \cite{zhang2017active,siddhant2018deep}.
All these gains would be highly desirable for AL.
Therefore improving the adoption of DNNs in AL is crucial, especially since the expected increases in performance could be either used to improve the classification results while using the same amount of data or to increase the efficiency of the labeling process by reducing the data and therefore the labeling efforts.
Based on these findings we identify research directions for future work in order to further advance (D)NN-based AL.

%% file: acknowledgements.tex
\section*{Acknowledgements}

We thank Gerhard Heyer for his valuable feedback on the manuscript, Lydia Müller for fruitful discussions about the taxonomy and advice thereon, and Janos Borst for sharing his thoughts on recent advances in language models.
This research was partially funded by the Development Bank of Saxony
(SAB) under project number 100335729.

%% file: tex/appendix.tex
\appendix
\section{Appendix}

\subsection{Datasets}

The following table provides additional information about the datasets which were referred to in Section \ref{subsect:active_tc}.

{
\setlength{\LTcapwidth}{\textwidth}
\begin{longtable}{l p{3.8cm} l p{4.1cm} r r}
  \hline
  \textbf{Id} & \textbf{Name} & \textbf{Type} & \textbf{Publication} & \textbf{\#Train} & \textbf{\#Test} \\
  \hline
  TQA & TREC QA & MC & \cite{li2002learning} & 5,500 & 500 \\
  CR & Customer Reviews & MC & \cite{hu2004mining} & \textsuperscript{*}{315} & - \\
  ACR & Additional Customer\newline Reviews & MC & \cite{ding2008a} & \textsuperscript{*}{325} & - \\
  MDS & Multi-Domain Sentiment & B & \cite{blitzer2007biographies} & \textsuperscript{**}{8,000} & - \\
  BAG & Blog Author Gender & B & \cite{mukherjee2010improving} & 3,100 & - \\
  G13 & Guardian 2013 & MC & \cite{belford2018topic} & 6,520 & -\\
  MR & Movie Reviews & B & \cite{pang2005seeing} & 10,662 & - \\
  MRL & Movie Reviews Long & B & \cite{pang2004sentimental}  & 2,000 & - \\
  MUR & Music Review & B & \cite{blitzer2007biographies} & 2,000 & - \\
  DR & Doctor Reviews & MC & \cite{wallace2014large-scale} & 58,110 & - \\
  SJ & Subjectivity & B & \cite{pang2004sentimental} & 10,000 & -\\
  20N & 20newsgroups & MC & \cite{joachims1997probabilistic} & \textsuperscript{***}18,846 & - \\
  R21 & Reuters-21578 & ML & \cite{lewis1997reuters} & 21578 & - \\
  RMA & Reuters ModApté & ML & \cite{apte1994towards} & 9,603 & 3,299\\
  RV2 & RCV1-V2 & ML & \cite{lewis2004rcv1} & 23,149 & 781,265 \\
  SPM & Spam & B & \cite{delany2005case-based} & 1,000 & - \\
  AGN & AG News & MC & \cite{gulli2005agnews}\newline\cite{zhang2016character-level} & 120,000 & 7,600 \\
  SGN & Sogou News & MC & \cite{wang2008automatic} & 450,000 & 60,000\\
  DBP & DBPedia & MC & \cite{zhang2016character-level} & 560,000 & 70,000\\
  YRP & Yelp Review Polarity & B & \cite{zhang2016character-level} & 560,000 & 38,000\\
  YRF & Yelp Review Full & MC & \cite{zhang2016character-level} & 650,000 & 50,000\\
  YAH & Yahoo! Answers & MC & \cite{zhang2016character-level} & 1,400,000 & 60,000\\
  ARP & Amazon Review Polarity & B & \cite{zhang2016character-level} & 3,600,000 & 40,000\\
  ARF & Amazon Review Full & MC & \cite{zhang2016character-level} & 3,000,000 & 650,000\\
  \hline
  \caption{A collection of widely-used text classification datasets.
  The column "Type" denotes the classification setting (B = binary, MC = multi-class, ML = multi-class multi-label). 
  The columns "\#Train" and "\#Test" show the size of the train and of the test set.
  In the case that no predefined splits were available "\#Train" represents the full dataset's size. 
  Each dataset was assigned a short id (first column), which we use in the paper for reference.\\
  \newline
  (*): documents, (**) labels reduced to positive/negative, (***) 20news-bydate with duplicates removed}
  \label{tab:datasets}
\end{longtable}
}